\documentclass[11pt,a4paper]{article}
\usepackage[hyperref]{acl2019}
\usepackage{array}
\usepackage{times}
\usepackage{latexsym}

\usepackage{url}

\usepackage{times}
\usepackage{latexsym}

\usepackage{times}
\usepackage{xcolor}
\usepackage{array}
\usepackage{tabularx}
\usepackage{graphicx}
\usepackage{amsmath}
\usepackage{amssymb}
\usepackage{bm}
\usepackage{booktabs,dcolumn}
\usepackage[utf8]{inputenc}

\newcolumntype{d}{D{.}{.}{2.5}}           % alignment on decimal marker
\newcolumntype{v}{>{\hsize=0.01\hsize}X}
\newcolumntype{t}{>{\hsize=0.04\hsize}X}
\newcolumntype{s}{>{\hsize=0.05\hsize}X}
\newcolumntype{b}{>{\hsize=0.4\hsize}X}

\newcolumntype{m}{>{\hsize=0.15\hsize}X}
\newcommand{\ra}[1]{\renewcommand{\arraystretch}{#1}}

\aclfinalcopy % Uncomment this line for the final submission
\title{Recognising Agreement and Disagreement between Stances with Reason Comparing Networks}

\author{Chang Xu, C\'ecile Paris, Surya Nepal, \and Ross Sparks\\
	CSIRO Data61, Australia\\
	\{Chang.Xu, Cecile.Paris, Surya.Nepal, Ross.Sparks\}@data61.csiro.au
}

\date{}

\begin{document}
\maketitle

\begin{abstract}
We identify agreement and disagreement between utterances that express stances towards a topic of discussion. 
Existing methods focus mainly on conversational settings, where dialogic features are used for (dis)agreement inference. 
We extend this scope and seek to detect stance (dis)agreement in a broader setting, where \textit{independent} stance-bearing utterances, which prevail in many stance corpora and real-world scenarios, are compared.
To cope with such non-dialogic utterances, we find that the reasons uttered to back up a specific stance can help predict stance (dis)agreements.
We propose a reason comparing network (RCN) to leverage reason information for stance comparison.
Empirical results on a well-known stance corpus show that our method can discover useful reason information, enabling it to outperform several baselines in stance (dis)agreement detection.
\end{abstract}

\section{Introduction}

Agreement and disagreement naturally arise when peoples' views, or ``stances'', on the same topics are exchanged.
Being able to identify the convergence and divergence of stances is valuable to various downstream applications, such as discovering subgroups in a discussion~\cite{hassan2012detecting,abu2012subgroup}, improving recognition of argumentative structure~\cite{lippi2016argumentation}, and bootstrapping stance classification with (dis)agreement side information~\cite{sridhar2014collective,ebrahimi2016weakly}. 
%help detect online disputes

Previous efforts on (dis)agreement detection are confined to the scenario of natural dialogues~\cite{misra2017topic,wang2014improving,sridhar2015joint,rosenthal2015couldn}, where dialogic structures are used to create a conversational context for (dis)agreement inference.
However, non-dialogic stance-bearing utterances are also very common in real-world scenarios.
For example, in social media, people can express stances autonomously, without the intention of initiating a discussion~\cite{mohammad2016semeval}.
There are also corpora built with articles containing many self-contained stance-bearing utterances~\cite{ferreira2016emergent,bar2017stance}.

Studying how to detect (dis)agreement between such \textit{independent} stance-bearing utterances has several benefits:
1) pairing these utterances can lead to a larger (dis)agreement corpus for training a potentially richer model for (dis)agreement detection;
2) the obtained pairs enable training a distance-based model for opinion clustering and subgroup mining;
3) it is applicable to the aforementioned non-dialogic stance corpora;
and 4) it encourages discovering useful signals for (dis)agreement detection beyond dialogic features (e.g., the reason information studied in this work).

In this work, we investigate how to detect (dis)agreement between a given pair of (presumably unrelated) stance-bearing utterances.
Table~\ref{tb:cp} shows an example where a decision is made on whether two utterances agree or disagree on a discussed topic.
This task, however, is more challenging, as the inference has to be made without using any contextual information (e.g., dialogic structures).
To address this issue, one may need to seek clues \textit{within} each of the compared utterances to construct appropriate contexts.

\begin{table}[!h]
	\footnotesize
	\centering
	\ra{1.1}
	\begin{tabular}{|p{0.205\textwidth}p{0.215\textwidth}|}
		\hline
		\multicolumn{2}{|c|}{Topic: \textbf{Gun Control}} \\
		\hline
		Utterance 1: & Utterance2: \\
		\textit{If guns are outlawed, only outlaws will have guns}. (Stance: Against) & \textit{Freedom to have a gun is same as freedom of speech}. (Stance: Against) \\
		\hline
		\multicolumn{2}{|c|}{Class Label: \textbf{Agree}}\\
		\hline
	\end{tabular}
	\caption{The task of detecting stance (dis)agreement between utterances towards a topic of discussion.}
	\label{tb:cp}
\end{table}

It has been observed that when expressing stances, people usually back up their stances with specific explanations or reasons~\cite{hasan2014you,boltuzic2014back}.
These reasons are informative about which stance is taken, because they give more details on how a stance is developed.
However, simply comparing the reasons may not be sufficient to predict stance (dis)agreement, as sometimes people can take the same stance but give different reasons (e.g., the points \textit{outlaws having guns} and \textit{freedom of speech} mentioned in Table~\ref{tb:cp}).
One way to address this problem is to make the reasons \textit{stance-comparable}, so that the reason comparison results can be stance-predictive.

In this paper, in order to leverage reason information for detecting stance (dis)agreement, we propose a \textit{reason embedding} approach, where the reasons are made stance-comparable by projecting them into a shared, embedded space.
In this space, ``stance-agreed'' reasons are close while ``stance-disagreed'' ones are distant.
For instance, the reason points \textit{outlaws having guns} and \textit{freedom of speech} in Table~\ref{tb:cp} would be near to each other in that space, as they are ``agreed'' on the same stance.
We learn such reason embedding by comparing the reasons using utterance-level (dis)agreement supervision, so that reasons supporting agreed (disagreed) stances would have similar (different) representations.
A reason comparing network (RCN) is designed to learn the reason embedding and predict stance (dis)agreement based on the embedded reasons.
Our method complements existing dialogic-based approaches by providing the embedded reasons as extra features.
We evaluate our method on a well-known stance corpus and show that it successfully aligns reasons with (dis)agreement signals and achieves state-of-the-art results in stance (dis)agreement detection.

\section{RCN: The Proposed Model}

\begin{figure}[b]
	\centering
	\includegraphics[width=\columnwidth]{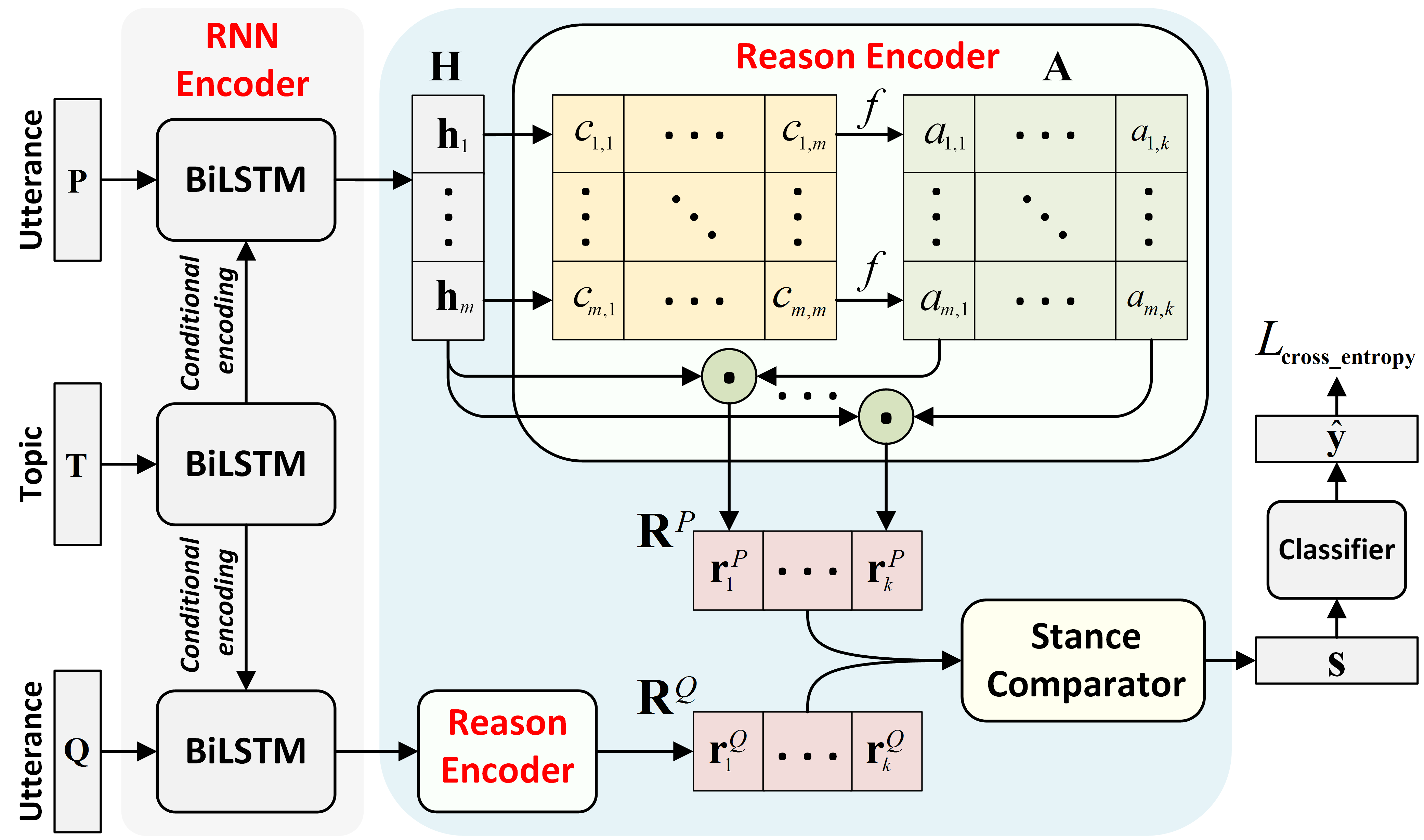}
	%\vspace{-1mm}
	\caption{The architecture of RCN.}
	%\vspace{-2mm}
	\label{fig:model}
\end{figure}

Figure~\ref{fig:model} illustrates the architecture of RCN. 
At a high level, RCN is a classification model that takes as input an utterance pair $(P,Q)$ and a topic $T$, and outputs a probability distribution $\textbf{y}$ over three classes $\{Agree,Disagree,Neither\}$ for stance comparison.
To embed reasons, RCN uses two identical sub-networks (each contains an RNN encoder and a reason encoder) with shared weights to extract reason information from the paired utterances and predict their stance (dis)agreement based on the reasons.
%The comparison result is then fed into a (dis)agreement classifier to produce the final prediction. 
%Next, we explain each component of RCN in details. 

\paragraph{RNN Encoder:}
In this module, we use RNNs to encode the input utterances.
We first use \textit{word embedding} to vectorise each word in the input utterance pair $(P,Q)$ and topic $T$, obtaining three sequences of word vectors $\textbf{P}$, $\textbf{Q}$, and $\textbf{T}$.
Then we use two BiLSTMs to encode the utterance and topic sequences, respectively.
Moreover, by following the work of \citet{augenstein2016stance}, we use \textit{conditional encoding} to capture the utterances' dependencies on the topic.
%, by initialising the hidden states of the utterance BiLSTM with the hidden states of the topic BiLSTM.
The output are two topic-encoded sequences produced by the utterance BiLSTM for $P$($Q$), denoted by $\textbf{H}^{P(Q)}=\{\textbf{h}_i^{P(Q)}\}_{i=1}^{|{P(Q)}|}\in\mathbb{R}^{|{P(Q)}|\times 2h}$, where $h$ is the hidden size of a unidirectional LSTM. 

%\begin{equation}
%\small
%\begin{split}
%&[\overrightarrow{\bm{h}}_i^T\,\overrightarrow{\bm{c}}_i^T]=\overrightarrow{\text{LSTM}}^T(\bm{t}_i, \overrightarrow{\bm{h}}_{i-1}^T, \overrightarrow{\bm{c}}_{i-1}^T)\quad i\in\{1,...,|T|\}\\
%&[\overleftarrow{\bm{h}}_i^T\,\overleftarrow{\bm{c}}_i^T]=\overleftarrow{\text{LSTM}}^T(\bm{t}_i, \overleftarrow{\bm{h}}_{i+1}^T, \overleftarrow{\bm{c}}_{i+1}^T)\quad i\in\{|T|,...,1\})
%\end{split}
%\end{equation}
%
%\begin{equation}
%\label{eq:bilstm_p}
%\small
%\begin{split}
%[\overrightarrow{\bm{h}}_1^P\,\overrightarrow{\bm{c}}_1^P]&=\overrightarrow{\text{LSTM}}^{PQ}(\bm{p}_1, \overrightarrow{\bm{h}}_{|T|}^T, \overrightarrow{\bm{c}}_{|T|}^T)\\
%[\overleftarrow{\bm{h}}_{|P|}^P\,\overleftarrow{\bm{c}}_{|P|}^P]&=\overleftarrow{\text{LSTM}}^{PQ}(\bm{p}_{|P|}, \overleftarrow{\bm{h}}_1^T, \overleftarrow{\bm{c}}_1^T)
%\end{split}
%\end{equation}

\paragraph{Reason Encoder:}
Then we extract reasons from the utterances, which is the main contribution of this work.
In particular, we focus on the \textit{major} reasons that most people are concerned with, which possess two properties: 1) they are focal points mentioned to support a specific stance; 2) they recur in multiple utterances.
With such properties, the extraction of these reasons can then be reduced to finding the \textit{recurring focal points} in all the input utterances.

To action on this insight, we take a weighting-based approach by learning a weighting matrix $\textbf{A}$ that captures the relatedness between each position in an utterance and each implied reason. 
For example, on utterance $P$ where we hypothesise $\kappa$ possible reasons, the weighting matrix is $\textbf{A}^{|P|\times \kappa}$, with each cell $\textbf{A}_{i,k}$ representing the relatedness between the $i$th position of $P$ and the $k$th reason.

To learn the weighting matrix $\textbf{A}$, we use \textit{self-attention}~\citep{cheng2016long} and develop a particular self-attention layer for implementing the above weighting scheme.
Meanwhile, the recurrence of a reason is also perceivable, as all utterances mentioning that reason are used to learn the self-attention layer.

Our particular self-attention applied on an utterance is designed as follows.
First, a pairwise relatedness score is computed between each pair of positions ($\textbf{h}_i$,$\textbf{h}_j$) with a bilinear transformation, $c_{i,j}=\text{tanh}(\textbf{h}_i^\top\textbf{W}^{(1)}\textbf{h}_j)$, where $\textbf{W}^{(1)}\in\mathbb{R}^{2h\times 2h}$ is a trainable parameter.
Next, for each position $\textbf{h}_i$, we convert its relatedness scores with all other positions into its overall relatedness scores with $\kappa$ possible reasons using a linear transformation $f$,
\begin{equation}
\label{eq:reason_score}
\small
e_{i, k}=f(\{c_{i,j}\}_{j=1}^{|P|\,\text{or}\,|Q|})=\sum_{j=1}^{|P|\,\text{or}\,|Q|}c_{i,j}\cdot W_{j,k}^{(2)}+b_k
\end{equation}
where $\textbf{W}^{(2)}=\{W_{j,k}^{(2)}\}\in\mathbb{R}^{|P\,\text{or}\,Q|\times\kappa}$ and $\textbf{b}\in\mathbb{R}^{\kappa}$ are trainable parameters.
The philosophy behind Eq.~\ref{eq:reason_score} is that the relatedness distribution $\{c_{i,j}\}$ of an utterance implies segments in it that are internally compatible, which may correspond to different focal points (reasons).
The transformation $f$ then learns the mapping between those two. 
Finally, we obtain the attention weight $\textbf{A}_{i,k}$ for each position $i$ on each reason $k$ by applying \textit{softmax} over all $e_{\ast,k}$s, 
\begin{equation}
\label{eq:attention}
\textbf{A}_{i,k}=\frac{\exp(e_{i,k})}{\sum_{j=1}^{|P|\,\text{or}\,|Q|} \exp(e_{j,k})}
\end{equation}

With $\textbf{A}$ obtained, we can compute an utterance's reason encoding as the sum of its RNN encoding $\{\textbf{h}_i\}$ weighted by $\textbf{A}$: $\textbf{r}_k=\sum_{i=1} a_{k,i}\textbf{h}_i$, where $\textbf{r}_k\in\mathbb{R}^{2h}$ is the encoding for the $k$th reason.
We use $\textbf{R}^{P(Q)}=[\textbf{r}_1^{P(Q)},...,\textbf{r}_{\kappa}^{P(Q)}]\in\mathbb{R}^{2h\times \kappa}$ to denote the \textit{reason matrix} for the utterance $P$($Q$).

It is worth noting that the above self-attention mechanism in our reason encoding can also be seen as a variant of \textit{multi-dimensional} self-attention, as we simultaneously learn multiple attention vectors for the different reasons implied in an utterance.
%Also note that the sentences can agree or disagree based on different reasons as long as such patterns contribute to predicting the consensus polarity (see examples in Section~\ref{sec:visual}).
%It is worth noting that for mixed-target datasets, it is natural for each target to have its own reason space.
%In our framework, this can be addressed by augmenting all the above parameters $\bm{W},\bm{b}$ with a new first dimension (the length equals to the total number of targets), and using a one-hot target-indicated vector $\bm{g}$ to select the respective sets of parameters via matrix multiplication $\bm{g}^\top\bm{W}$ and $\bm{g}^\top\bm{b}$.
%Also note that our reason-aware mechanism is not intended to extract the exact reasons behind stance. 
%Instead, it aims to emphasize the parts of an utterance that could be closely related to the author's reasoning behind.

\paragraph{Stance Comparator:}
Now we compare the stances of $P$ and $Q$ based on their reason matrices.
Since we have captured multiple reasons in each utterance, all the differences between their reasons must be considered. 
We thus take a \textit{reason-wise} comparing approach, where every possible pair of reasons between $P$ and $Q$ is compared.
We employ two widely used operations for the comparison, i.e., \textit{multiplication}: $\textbf{s}_{i,j}^{\textbf{mul}}=\textbf{r}_i^P\odot\textbf{r}_j^Q$ and \textit{subtraction}: $\textbf{s}_{i,j}^{\textbf{sub}}=(\textbf{r}_i^P-\textbf{r}_j^Q)\odot(\textbf{r}_i^P-\textbf{r}_j^Q)$, where $\odot$ denote element-wise multiplication.
%These two operations \textbf{mul} and \textbf{sub} closely follow the well-known cosine similarity and Euclidean distance, except they do not reduce to a single summarised value.
%In this way, they can discover fine-grained differences between the dimensions of the compared reasons.
We then aggregate all the differences resulting from each operation into a single vector, by using a global max-pooling to signal the \textit{largest} difference with respect to an operation,
\begin{equation}
\small
\begin{split}
\textbf{s}^{\textbf{mul}}&=global\_max\_pooling(\{\textbf{s}_{i,j}^{\textbf{mul}}|i,j\in[1,\kappa]\})\\
\textbf{s}^{\textbf{sub}}&=global\_max\_pooling(\{\textbf{s}_{i,j}^{\textbf{sub}}|i,j\in[1,\kappa]\})
\end{split}
\end{equation}
The concatenation of the two difference vectors $\textbf{s}=[\textbf{s}^{\textbf{mul}};\textbf{s}^{\textbf{sub}}]$ forms the output of this module.
%\begin{equation}
%\small
%\textbf{s}^{\textbf{mul}}=\sum_{i,j}\textbf{s}_{i,j}^{\textbf{mul}}\,\,\,\text{and}\,\,\,\textbf{s}^{\textbf{sub}%}=\sum_{i,j}\textbf{s}_{i,j}^{\textbf{sub}}\quad i,j\in\{0,...,r-1\}
%\end{equation}

\paragraph{(Dis)agreement Classifier:}
Finally, a classifier is deployed to produce the (dis)agreement class probability $\hat{\textbf{y}}=\{\hat{y}_1,\hat{y}_2,\hat{y}_3\}$ based on the comparison result $\textbf{s}$, which consists of a two-layer feed-forward network followed by a softmax layer, $\hat{\textbf{y}}=\text{softmax}(\text{FeedForward}(\textbf{s}))$.

\paragraph{Optimisation:}
To train our model, we use the multi-class cross-entropy loss,
\begin{equation}
\label{eq:obj}
%\small
\mathcal{L}(\theta)=-\sum_i^N\sum_j^3 y_j^{(i)} \log\hat{y}_j^{(i)}+\lambda\sum_{\theta\in\Theta} \theta^2
\end{equation}
where $N$ is the size of training set, $y\in\{Agree,Disagree,Neither\}$ is the ground-truth label indicator for each class, and $\hat{y}$ is the predicted class probability.
$\lambda$ is the coefficient for $L_2$-regularisation.
$\Theta$ denotes the set of all trainable parameters in our model.

Minimising Eq.~\ref{eq:obj} encourages the comparison results between the extracted reasons from $P$ and $Q$ to be stance-predictive.

\section{Related Work}

Our work is mostly related to the task of detecting agreement and disagreement in online discussions.
Recent studies have mainly focused on classifying (dis)agreement in dialogues~\cite{abbott2011can,wang2014improving,misra2017topic,allen2014detecting}.
In these studies, various features (e.g., structural, linguistic) and/or specialised lexicons are proposed to recognise (dis)agreement in different dialogic scenarios.
In contrast, we detect stance (dis)agreement between independent utterances where dialogic features are absent.
%\citet{misra2017topic} proposed a set of topic-independent features motivated from dialogic theories for recognising rejection/disagreement in online debate forums.
%\citet{wang2014improving} constructed a specialised sentiment lexicon for (dis)agreement detection using conversational data from Wikipedia.
%\citet{allen2014detecting} relied on extracting a discourse tree structure from a conversation for feature building.
%Earlier efforts on this have examined the usefulness of various features (e.g., structural, linguistic) in recognising (dis)agreement in different dialogic scenarios (e.g., quote/response pairs in online arguments~\citep{abbott2011can} and broadcast conversations~\citep{wang2011detection}).

Stance classification has recently received much attention in the opinion mining community.
Different approaches have been proposed to classify stances of individual utterances in ideological forums~\citep{murakami2010support,somasundaran2010recognizing,gottopati2013learning,qiu2015modeling} and social media~\citep{augenstein2016stance,du2017stance,mohammad2017stance}.
%Many approaches to identifying stances exploit features extracted from the text of an utterance alone.
%More recent methods have realised the influence of text describing the discussion topic and included the topic sequence for model building~\cite{augenstein2016stance,du2017stance}.
In our work, we classify (dis)agreement relationships between a pair of stance-bearing utterances.

%There has also been research on joint modelling of stance and (dis)agreement between stances~\cite{sridhar2014collective,sridhar2015joint,ebrahimi2016weakly}.
%For instance, reply activities in online forums such as rebuttal or reply links among utterances are treated as (dis)agreement signals and used to bootstrap stance classification~\cite{walker2012stance,sridhar2014collective}.
%Social network structure and user interactions are used to supervised stance classification~\cite{dong2017weakly}.
%In this work, the reason embedding learned by our model could provide another feature dimension for building stance classification models.

Reason information has been found useful in argumentation mining~\cite{lippi2016argumentation}, where studies leverage stance and reason signals for various argumentation tasks~\cite{hasan2014you,boltuzic2014back,sobhani2015argumentation}.
We study how to exploit the reason information to better understand the stance, thus addressing a different task.
%Moreover, the reasons used in the above studies are pre-defined and hard-coded, which, in contrast, are automatically learned in our method.

Our work is also related to the tasks on textual relationship inference, such as textual entailment~\cite{snli:emnlp2015}, paraphrase detection~\cite{yin2015convolutional}, and question answering~\cite{wang2016inner}.
Unlike the textual relationships addressed in those tasks, the relationships between utterances expressing stances do not necessarily contain any rephrasing or entailing semantics, but they do carry discourse signals (e.g., reasons) related to stance expressing.

\section{Experiments}

%In this section, we perform a series of qualitative and quantitative experiments to evaluate the effectiveness of RCN in detecting agreement and disagreement between random utterances.
\subsection{Setup}
\paragraph{Dataset:}
The evaluation of our model requires a corpus of agreed/disagreed utterance pairs.
For this, we adapted a popular corpus for stance detection, i.e., a collection of tweets expressing stances from SemEval-2016 Task 6.
It contains tweets with stance labels (\textit{Favour}, \textit{Against}, and \textit{None}) on five topics, i.e., \textit{Climate Change is a Real Concern} (CC), \textit{Hillary Clinton} (HC), \textit{Feminist Movement} (FM), \textit{Atheism} (AT), and \textit{Legalization of Abortion} (LA).
We generated utterance pairs by randomly sampling from those tweets as follows:
\textit{Agreement} samples: 20k pairs labelled as (\textit{Favour}, \textit{Favour}) or (\textit{Against}, \textit{Against});
\textit{Disagreement} samples: 20k pairs as (\textit{Favour}, \textit{Against}), (\textit{Favour}, \textit{None}), or (\textit{Against}, \textit{None});
\textit{Unknown} samples: 10k pairs as (\textit{None}, \textit{None})\footnote{Fewer \textit{unknown} pairs being sampled is due to the inherently fewer \textit{none}-stance tweets in the original corpus.}.
%We use the rules mentioned above to create our corpora of agreement/disagreement pairs on these five topics.
%For each topic, we create 10 stratified folds by randomly sampling from the entire population ten times.
%\footnote{The sampling has considered the original class distributions in each corpus by weighting each pair according to its class proportion.}
%Table~\ref{tb:datasets} shows the class distribution of a typical fold. 
%\begin{table}
%	\small
%	 \begin{tabular}{|c|c|c|c|}
%		\hline 
%		  & Agreement & Disagreement & Unknown \\ 
%		\hline 
%		 Train & 20,000 & 20,000 & 10,000 \\ 
%		\hline 
%		Validation & 2,000 & 2,000 & 1,000 \\ 
%		\hline 
%		Test & 2,000 & 2,000 & 1,000 \\ 
%		\hline 
%	\end{tabular} 
%	\caption{The class distribution of our corpus.}
%	\label{tb:datasets}
%\end{table}

%We also create an ALL corpus by combining the above five corpora together.
%The ALL corpus consists of 250k pairs for training, 25k for validation, and 25k for testing.

\paragraph{Baselines:}
We compared our method with the following baselines: 1) BiLSTM: a base model for our task, where only the RNN encoder is used to encode the input;
%The encoded representations are aggregated into a fixed-length vector, which is then fed into the (dis)agreement classifier to generate the prediction.
2) DeAT~\cite{parikh2016decomposable}: a popular attention-based models for natural language inference.
%It utilises attention to decompose a text sequence into different aligned parts and then compare the parts between a pair of texts;
3) BiMPM~\cite{wang2017bilateral}: a more recent natural language inference model where two pieces of text are matched from multiple perspectives based on pooling and attention.
%It achieves state-of-the-art results on several text matching tasks such as textual entailment, paraphrase detection, and answer sentence selection.

%In order for all the baselines to account for the specific input of the \textit{topic} sequence $T$ to our task, we add the conditional encoding component~\citep{augenstein2016stance} to each baseline to enable the modelling of the dependency of the utterances on the topic, as we do in RCN.

%\subsection{Metric}
%We use macro \textit{F1-score} to evaluate the classification performance of each compared method.
%In particular, we have $F_{avg}=\frac{1}{3}(F_{agree}+F_{disagree}+F_{unknown})$, where $F=\frac{2\times Precision\times Recall}{Precision+Recall}$.
%For each topic, we report the results by averaging over all folds.
%The two-tailed t-test is used for the significance testing, yielding standard errors on the averaged results.

\paragraph{Training details:}
%We trained separate models on different topics.
An 80\%/10\%/10\% split was used for training, validation and test sets.
All hyper-parameters were tuned on the validation set.
The word embeddings were statically set with the 200-dimensional GloVe word vectors pre-trained on the 27B Twitter corpus.
%Stop words are removed in the preprocessing.
The hidden sizes of LSTM and FeedForward layers were set to 100.
A light dropout (0.2) was applied to DeAT and heavy (0.8) to the rest.
%The cross-entropy loss was used for training, 
ADAM was used as the optimiser and learning rate was set to $10^{-4}$.
%All models were trained via batches of size 256.
Early stopping was applied with the patience value set to 7.
%Our experiments run on a Tesla P100 with 16GB memory.

\begin{table}[!t]
	\small
	\ra{1}
	\centering
	\begin{tabular}{ccccl}
		\toprule
		Topic & {BiLSTM} & {DeAT} & {BiMPM} & {RCN} (Our)\\
		\midrule
		CC        & 68.1$\pm$0.6    & 70.9$\pm$0.7   &  71.5$\pm$0.6   & \textbf{73.0$\pm$0.5}$^{*}$   \\
		HC        & 52.5$\pm$0.6    & 56.9$\pm$0.4  & 56.4$\pm$0.7   & \textbf{58.6$\pm$0.4}$^{**}$   \\
		FM        & 58.3$\pm$0.6    & 60.6$\pm$0,7  & 59.8$\pm$0.7   & \textbf{64.4$\pm$0.5}$^{**}$   \\
		AT        & 67.5$\pm$0.4    & 69.5$\pm$0.5  & 70.3$\pm$0.6    & \textbf{72.2$\pm$0.4}$^{*}$   \\
		LA        & 61.3$\pm$0.3    & 63.2$\pm$0.6   & 62.4$\pm$0.4   & \textbf{64.5$\pm$0.4}$^{**}$   \\
		%ALL      & 58.7$\pm$0.3    & 60.7$\pm$0.5  & 62.1$\pm$0.2   & \textbf{65.0$\pm$0.2}   \\
		\midrule[\heavyrulewidth]
		\multicolumn{5}{l}{Two tailed t-test: $^{**}\ p<0.01$; $^{*}\ p<0.05$}		\\
	\end{tabular}
	%\vspace{-2mm}
	\caption{Classification performance of the compared methods on various topics, measured by the averaged macro F1-score over ten runs on the test data.}
	%\vspace{-6mm}
	\label{tb:f-score}
\end{table}

\begin{table*}[!h]
	\ra{1}
	\scriptsize
	\centering
	\begin{tabularx}{\textwidth}{vtsbb}
		\toprule
		\textbf{ID} & \textbf{Label} & \textbf{Topic}  & \textbf{Tweet 1} & \textbf{Tweet 2}  \\
		\midrule
		{1} & Agree &  HC \par(A, A)   &  \textbf{Reason 1:}\colorbox{red!0}{\strut @HillaryClinton}\colorbox{red!0}{\strut is}\colorbox{red!0}{\strut a}\colorbox{red!43}{\strut liar}\colorbox{red!0}{\strut \&}\colorbox{red!100}{\strut corrupt}.\colorbox{red!0}{\strut Period.}\par\colorbox{red!0}{\strut End}\colorbox{red!0}{\strut of}\colorbox{red!0}{\strut story.} & \textbf{Reason 1:}\colorbox{red!2}{\strut @HillaryClinton}\colorbox{red!100}{\strut lies}\colorbox{red!0}{\strut just}\colorbox{red!0}{\strut for}\colorbox{red!0}{\strut the}\colorbox{red!11}{\strut fun}\colorbox{red!0}{\strut of}\colorbox{red!0}{\strut it,}\colorbox{red!0}{\strut its}\par\colorbox{red!0}{\strut CRAZY!!!!!} \\
		\midrule
		{2} &  Disagree &  LA \par (F, A) & \textbf{Reason 1:}\colorbox{red!0}{\strut I}\colorbox{red!0}{\strut would}\colorbox{red!15}{\strut never}\colorbox{red!11}{\strut expect}\colorbox{red!0}{\strut an}\colorbox{red!9}{\strut 11}\colorbox{red!17}{\strut year}\colorbox{red!31}{\strut old}\colorbox{red!100}{\strut girl}\colorbox{red!0}{\strut to}\colorbox{red!0}{\strut have}\par\colorbox{red!0}{\strut to}\colorbox{red!83}{\strut carry}\colorbox{red!0}{\strut a}\colorbox{red!84}{\strut pregnancy}\colorbox{red!0}{\strut to}\colorbox{red!0}{\strut term} &  \textbf{Reason 1:}\colorbox{red!2}{\strut Actually,}\colorbox{red!54}{\strut child}\colorbox{red!100}{\strut murder}\colorbox{red!0}{\strut is}\colorbox{red!13}{\strut far}\colorbox{red!15}{\strut worse}\colorbox{red!0}{\strut these}\colorbox{red!2}{\strut days.}\par\colorbox{red!0}{\strut We}\colorbox{red!14}{\strut live}\colorbox{red!0}{\strut in}\colorbox{red!0}{\strut more}\colorbox{red!11}{\strut savage}\colorbox{red!0}{\strut times.}\\
		\midrule
		{3} &  Agree & CC \par (F, F) & \textbf{Reason 1:}\colorbox{red!0}{\strut Living}\colorbox{red!0}{\strut an}\colorbox{red!100}{\strut unexamined}\colorbox{red!1}{\strut \#life}\colorbox{red!1}{\strut may}\colorbox{red!0}{\strut be}\colorbox{red!0}{\strut easier}\colorbox{red!0}{\strut but}\par\colorbox{red!0}{\strut leads}\colorbox{red!0}{\strut to}\colorbox{red!0}{\strut disastrous}\colorbox{red!0}{\strut consequences.} & \textbf{Reason 1:}\colorbox{red!0}{\strut There's}\colorbox{red!0}{\strut no}\colorbox{red!0}{\strut more}\colorbox{red!9}{\strut normal}\colorbox{red!100}{\strut rains}\colorbox{red!2}{\strut anymore.}\colorbox{red!6}{\strut Always}\par\colorbox{red!16}{\strut storms},\colorbox{red!4}{\strut heavy}\colorbox{red!0}{\strut and}\colorbox{red!15}{\strut flooding}.  \\ 
		&  &  & \textbf{Reason 2:}\colorbox{red!2}{\strut Living}\colorbox{red!0}{\strut an}\colorbox{red!0}{\strut unexamined}\colorbox{red!19}{\strut \#life}\colorbox{red!26}{\strut may}\colorbox{red!0}{\strut be}\colorbox{red!13}{\strut easier}\colorbox{red!0}{\strut but}\par\colorbox{red!7}{\strut leads}\colorbox{red!0}{\strut to}\colorbox{red!50}{\strut disastrous}\colorbox{red!100}{\strut consequences}. & \textbf{Reason 2:}\colorbox{red!0}{\strut There's}\colorbox{red!0}{\strut no}\colorbox{red!0}{\strut more}\colorbox{red!0}{\strut normal}\colorbox{red!0}{\strut rains}\colorbox{red!1}{\strut anymore.}\colorbox{red!89}{\strut Always}\par\colorbox{red!90}{\strut storms},\colorbox{red!1}{\strut heavy}\colorbox{red!0}{\strut and}\colorbox{red!1}{\strut flooding.} \\
		\midrule[\heavyrulewidth]
	\end{tabularx}
	\vspace{-3mm}
	\caption{The heatmaps of the attention weights assigned by the attention layer in the reason encoder to three tweet-pair examples. 
		In each example, we show the text of each tweet, the topic, the correct (dis)agreement label, and the stance of each tweet (F: Favour, A: Against).}
	%\vspace{-2mm}
	\label{tb:visual}
\end{table*}

\begin{table}[!h]
	%\vspace{-2mm}
	\scriptsize
	%\small
	\centering
	\ra{1.4}
	\begin{tabular}{|@{}c@{}|p{0.81\columnwidth}|}
		\hline
		\footnotesize \textbf{Topic} & \footnotesize \textbf{Top reason words ranked by attention weights} \\ 
		\hline
		\scriptsize CC & environment, climate, sustainability, safety, economy, community, good, kill, drought, insane, proud, co2, coal, clean, green\\
		\hline
		\scriptsize HC  & candidate, freedom, liberal, disappointed, greed, democrat, cheat, illegal, best, economy, war, american, republican, cutest\\
		\hline
		\scriptsize FM  & women, husband, divorce, girlfriend, adorable, ignorant, rights, behaved, marriage, infanticide, gender, queen, child, equality\\
		\hline
		\scriptsize AT  & fear, evil, jesus, human, truest, god, pray, belief, religion, ancient, tribulation, love, sovereign, church, secular, ignorance\\
		\hline
		\scriptsize LA  & pregnant, abortionist, murder, accidental, right, fertility, justice, illegal, democrat, marriage, government, motherhood, freedom\\
		\hline
	\end{tabular}
	\vspace{-1mm}
	\caption{The reason words learned on various topics.}
	\vspace{-3mm}
	\label{tb:pool}
\end{table}

\subsection{Results}

Table~\ref{tb:f-score} shows the results of our method and all the baselines on tasks with different topics.
We can first observe that the proposed RCN consistently outperformed all the baselines across all topics.
Despite being modest, all the improvements of RCN over the baselines are statistically significant at $p<0.05$ with a two-tailed t-test.
Among these methods, BiLSTM performed the worst, showing that only using the RNN encoder for sequence encoding is not sufficient for obtaining optimal results.
DeAT and BiMPM performed similarly well; both used attention to compare the utterances at a fine-grained level, resulting in a 2$\sim$5\% boost over BiLSTM.
Finally, RCN performed the best, with relative improvements from 2.1\% to 10.4\% over the second best.
As all the compared methods shared the same RNN encoding layers, that RCN performed empirically the best demonstrates the efficacy of its unique reason encoder and stance comparator in boosting performance.

%\subsubsection{Full spectrum evaluation}
%The last rows of the two Tables show the results on the ALL dataset where all topics are mixed together.
%Again, we observe a similar trend in performance.
%However, several differences are noticed.
%First, DeAT performs slightly better than BiLSTM, but worse than BiMPM by a larger margin than before.
%This could be due to the fact that DeAT employs a similar but simpler attention mechanism than BiMPM; the latter adopts a multi-perspective matching to enhance modelling, which introduces more dimensions for encoding the relationships between utterances, thus resulting in a moderate boost over DeAT.
%Finally, we notice that RCN dominates the performance, with even larger improvements in general than those in the previous per-topic evaluation.
%This suggests that RCN is more effective in handling the interplay of different topics in a mixed-topic setting. 

\subsection{Analysis}
\label{sec:visual}
In this section, we study what has been learned in the reason encoder of RCN.
In particular, we show the attentive activations in the reason encoder (i.e., $\textbf{A}$ in Eq.~\ref{eq:attention}), and see if reason-related contents could draw more attention from RCN.

\paragraph{Visualising attention signals in tweets:}
Table~\ref{tb:visual} shows the attention activations on three examples of tweet pairs chosen from our test set.
For the first two, we set the number of reasons to be attended to as one.
It can be seen that the parts of the tweets that received large attention weights (the highlighted words in Table~\ref{tb:visual}) were quite relevant to the respective topics; \textit{liar}, \textit{corrupt}, and \textit{lie} are words appearing in news about \textit{Hillary Clinton}; \textit{girl}, \textit{pregnancy}, and \textit{murder} are common words in the text about \textit{Legalisation of Abortion}.
Also, most of the highlighted words have concrete meanings and are useful to understand why the stances were taken.
The last row shows a case when two reasons had been attended to.
We observe a similar trend as before that the highlighted contents were topic-specific and stance-revealing.
Moreover, since one more reason dimension was added to be inferred in this case, RCN was able to focus on different parts of a tweet corresponding to the two reasons.
%For example, in Tweet 1 of example 3, \textit{unexamined life} and \textit{disastrous consequences} are the two points made in the text and are correctly highlighted by the learned reason vectors.
%Finally, we notice that some sentiment words are also emphasised (mildly) by our model, such as \textit{never expect} and \textit{far worse} in example 2, which demonstrates the role of sentiment words in stance expressions, with similar results also found in~\cite{mohammad2017stance}.

\paragraph{Visualising learned reasons:}
We also visualised the reasons learned by our model, represented as the words assigned with the largest attention weights in our results (i.e., 1.0).
Table~\ref{tb:pool} shows samples of such reason words. 
We see that the reason words have strong correlations with the respective topics, and, more importantly, they reflect different reason aspects regarding a topic, such as \textit{economy} vs. \textit{community} on \textit{Climate Change is a Real Concern} and \textit{culture} vs. \textit{justice} on \textit{Legalisation of Abortion}.
%Finally, we can also see some sentiment words (\textit{good}, \textit{insane}) that contribute to predicting the (dis)agreement labels.

In summary, both the visualisations in Table~\ref{tb:visual} and \ref{tb:pool} show that the attention mechanism employed by RCN is effective in finding different reason aspects that contribute to stance comparison.

%\section{Conclusion and Future Work}
\section{Conclusion and Future Work}
In this paper, we identify (dis)agreement between stances expressed in paired utterances.
%We extend the task scope from traditional dialogic settings to comparing stances among arbitrary utterances.
We exploit the reasons behind the stances and propose a reason comparing network (RCN) to capture the reason information to infer the stance (dis)agreement.
A quantitative analysis shows the effectiveness of RCN in recognising stance (dis)agreement on various topics.
A visualisation analysis further illustrates the ability of RCN to discover useful reason aspects for the stance comparison.

In the future, this work can be progressed in several ways.
First, it is necessary to evaluate our model on more stance data with different linguistic properties (e.g., the much longer and richer stance utterances in posts or articles).
Second, it is important to show how the learned embedded reasons can help downstream applications such as stance detection.
Finally, it would be insightful to further visualise the reasons in the embedded space with more advanced visualisation tools.

\section*{Acknowledgements}
We thank all anonymous reviewers for their constructive comments. 
We would also like to thank Xiang Dai for his helpful comments on drafts of this paper.
\bibliography{naaclhlt2019}
\bibliographystyle{acl_natbib}

\end{document}